\documentclass{article}

\usepackage{arxiv}

\usepackage[utf8]{inputenc} 
\usepackage[T1]{fontenc}    
\usepackage{hyperref}       
\usepackage{url}            
\usepackage{booktabs}       
\usepackage{amsfonts}       
\usepackage{nicefrac}       
\usepackage{microtype}      
\usepackage{lipsum}		
\usepackage{graphicx}
\usepackage{natbib}
\usepackage{doi}
\usepackage{amsmath}
\usepackage{color}
\usepackage{float}
\usepackage{algorithm2e}
\usepackage{algorithmic}
\usepackage{subfigure}
\usepackage{mathtools}
\RestyleAlgo{ruled} 

\title{Reinforced Inverse Scattering}

\author{ {\hspace{1mm}Hanyang Jiang}\\
	School of Industrial and Systems Engineering\\
	Georgia Institute of Technology\\
	\texttt{scottjhy@gatech.edu} 
	\And
	{\hspace{1mm}Yuehaw Khoo} \\
	Department of Statistics\\
	University of Chicago\\
	\texttt{ykhoo@uchicago.edu} 
	\And
	{\hspace{1mm}Haizhao Yang} \\
	Department of Mathematics\\
 Department of Computer Science\\
	University of Maryland College Park\\
	\texttt{hzyang@umd.edu} 
}

\renewcommand{\shorttitle}{Reinforced Inverse Scattering}

\hypersetup{
pdftitle={A template for the arxiv style},
pdfsubject={q-bio.NC, q-bio.QM},
pdfauthor={Hanyang Jiang, Yuehaw Khoo, Haizhao Yang},
pdfkeywords={Inverse Scattering, Reinforcement Learning},
}

\begin{document}
\maketitle

\begin{abstract}
Inverse wave scattering aims at determining the properties of an object using data on how the object scatters incoming waves. To gather this information, sensors are positioned to transmit and receive waves. The effectiveness of reconstructing scatterer properties depends significantly on the placement of these sensors and the frequencies of the incident waves. This paper presents a reinforcement learning framework to enhance user-specified strategies by adaptively determining optimal sensor locations and wave frequencies tailored to different scatterers. This approach leads to a notable enhancement in reconstruction quality, even with limited imaging resources. Extensive numerical results will be provided to demonstrate that the usage of a reinforcement learning framework is beneficial.
\end{abstract}


\section{Introduction}
\label{sec1}
Nowadays, artificial intelligence (AI) has fundamentally transformed a vast array of industries. AI harnesses the power of computers and machines to replicate the problem-solving and decision-making capabilities of the human mind. In many fields, AI has already achieved performance levels comparable to those of human experts, as seen in image recognition \citep{ir}, the game of Go \citep{go}, Starcraft \citep{star}, and voice generation \citep{wavenet}. Recently, AI has found applications in numerous scientific domains, including protein structure prediction \citep{fold}, climate forecasting \citep{clim}, and astronomical pattern recognition \citep{ast}. These remarkable successes have spurred interest in exploring AI's potential in various scientific research areas.

This paper introduces reinforcement learning (RL) to address inverse scattering problems. Specifically, a reinforcement learning framework is designed to autonomously learn and determine sensor positions and wave frequencies that adapt intelligently to different scatterers. Given an initial strategy, this approach leads to significant enhancements in reconstruction quality while making efficient use of imaging resources.

The inverse scattering problem involves the reconstruction or recovery of the physical and/or geometric properties of an object based on measured data. The information of interest typically includes the distribution of dielectric constants and the shape or structure of the object. The probing radiation can take the form of electromagnetic waves (e.g., microwave, optical wave, and X-ray), acoustic waves, or other waveforms. Inverse scattering holds great significance when detailed information about an object's structure and composition is needed. It finds broad applications in nondestructive evaluation, medical imaging, remote sensing, seismic exploration, target identification, geophysics, optics, atmospheric sciences, and various other fields \citep{app1,app2,app3,app4,app5,s1}.

We focus on the two-dimensional time-harmonic acoustic inverse scattering problem as a proof of concept for the reinforcement learning framework. In a compact domain $\Omega$ of interest, the inhomogeneous media scattering problem at a fixed wavenumber $k$ can be modeled by the following equation:
\begin{equation}
-\Delta u_{\text{out}} + k^2 \eta u_{\text{out}} = -k^2 \eta u_{\text{in}},
\label{holm}
\end{equation}
where $\eta$ is an unknown scatterer, $u_{\text{in}}$ represents the incoming wave field, and $u_{\text{out}}$ is the scattered wave field. The aim of the inverse problem is to recover the unknown $\eta$ given $u_{\text{in}}$, $k$, and the observed data measurements of $u_{\text{out}}$ denoted as $d$. Sensors are placed to transmit and receive scattered waves. The intrinsic properties of scatterers are hidden in the measurements $d$. A corresponding forward problem aims at computing $d$ from a given $\eta$. Both problems are computationally challenging. It is difficult to get a numerical solution for the inverse problem because of the nonlinearity of reconstructing $\eta$. Traditionally, there are several numerical methods attempting to resolve this inverse problem, which can be mainly divided into two types: nonlinear-optimization-based iterative methods \cite{o1,o2,o3} and imaging-based direct methods \cite{i1,i2,i3}. Recently, deep learning has been introduced to solve inverse scattering problems with new developments \cite{d1,d2,d3,rec,add}.

Inverse scattering problems are ill-posed when the incident wave has only one frequency due to the lack of stability \cite{ill1}. Minor variations in measured data may lead to significant errors in the reconstruction \cite{sta1,sta2}. To address this issue, various approaches have been proposed. One direction involves regularization methods under single-frequency data \cite{single1,single2}, which aim to enhance reconstruction efficiency and stability. Another approach is to use multi-frequency data in time-harmonic scattering problems \cite{m1,m2}. It has been shown that the inverse problem is uniquely solvable and Lipschitz stable when the highest wavenumber exceeds a certain threshold. However, the nonlinear equation becomes more oscillatory at higher frequencies and contains numerous local minima.

To overcome these challenges, \cite{chen1997inverse} developed a recursive-linearization-based algorithm that leverages multi-frequency data to form a continuation procedure, combining the advantages of low and high frequencies. Specifically, it solves the essentially linear equation at the lowest wavenumber and then uses the solution to incrementally linearize the equation for higher frequencies. This approach has laid the groundwork for further innovations in the field.

Significant progress has also been made in other directions for inverse scattering. For instance, single-frequency algorithms can be extended to multi-frequency versions, as demonstrated by \cite{mr1}. Moreover, \cite{mr3} introduced a novel Fourier method that directly reconstructs acoustic sources from multi-frequency measurements, avoiding the computational expense associated with iterative methods. Following the line of recursive-linearization-based algorithms, \cite{bao2007inverse} combined this approach with a direct imaging method as a warm-start technique for recovering obstacles using an inverse volume solver. Additionally, \cite{sini2012inverse, borges2015inverse} applied the recursive linearization algorithm to recover sound-soft obstacle shapes, achieving notable success.

Further advancements were made by \cite{borges2017high}, who achieved remarkably high-resolution reconstructions of smooth functions using recursive linearization, marking a significant milestone in 2D problems. In a recent study, \cite{borges2023robustness} highlighted the robustness of the inverse volume solver approach for obstacle recovery, demonstrating its superiority over methods focused solely on shape recovery. These developments underscore the continuous evolution of techniques to improve the accuracy and efficiency of solving inverse scattering problems.

Current literature focuses on computational algorithms that use fixed sensor positions and frequencies. Motivated by the numerical challenges and the aforementioned works, this paper proposes a reinforcement learning framework to select scatterer-dependent sensor locations and multiple frequencies to improve image reconstruction stability and quality for precision imaging. Previously, reinforcement learning has been applied to medical imaging \cite{ct,med,med2}, a related numerical problem to inverse scattering. Our algorithm is mainly inspired by the work \cite{ct}, where reinforcement learning is used to learn sensor locations and X-ray doses in CT imaging. It is important to note that the reconstruction problem in CT imaging is linear, while the one in inverse scattering is nonlinear and, hence, more challenging. Additionally, the goal in CT imaging is to optimize sensor locations by balancing sensing safety and reconstruction quality, whereas the goal in inverse scattering is to balance sensing expense and reconstruction quality, resulting in a different learning target in this paper than in \cite{ct}. Finally, we develop a new reinforcement learning framework that optimizes not only sensor locations but also incident wave frequencies. This paper focuses on the case of relatively weak scatterers as a proof of concept, while still maintaining the nonlinear nature of the forward problem. Extensive numerical results will be provided to demonstrate that our algorithm is capable of improving a user-specified strategy.

The rest of the paper is organized as follows. In Section \ref{sec2}, the inverse scattering problem is introduced. In Section \ref{sec3}, we explain the proposed reinforcement learning framework, the training and the testing procedure. In Section \ref{sec4}, numerical results are provided to demonstrate the effectiveness of the proposed framework. In Section \ref{sec5}, we conclude this paper with a short discussion.

\begin{figure}[h!]
\centering
\includegraphics[scale=0.15]{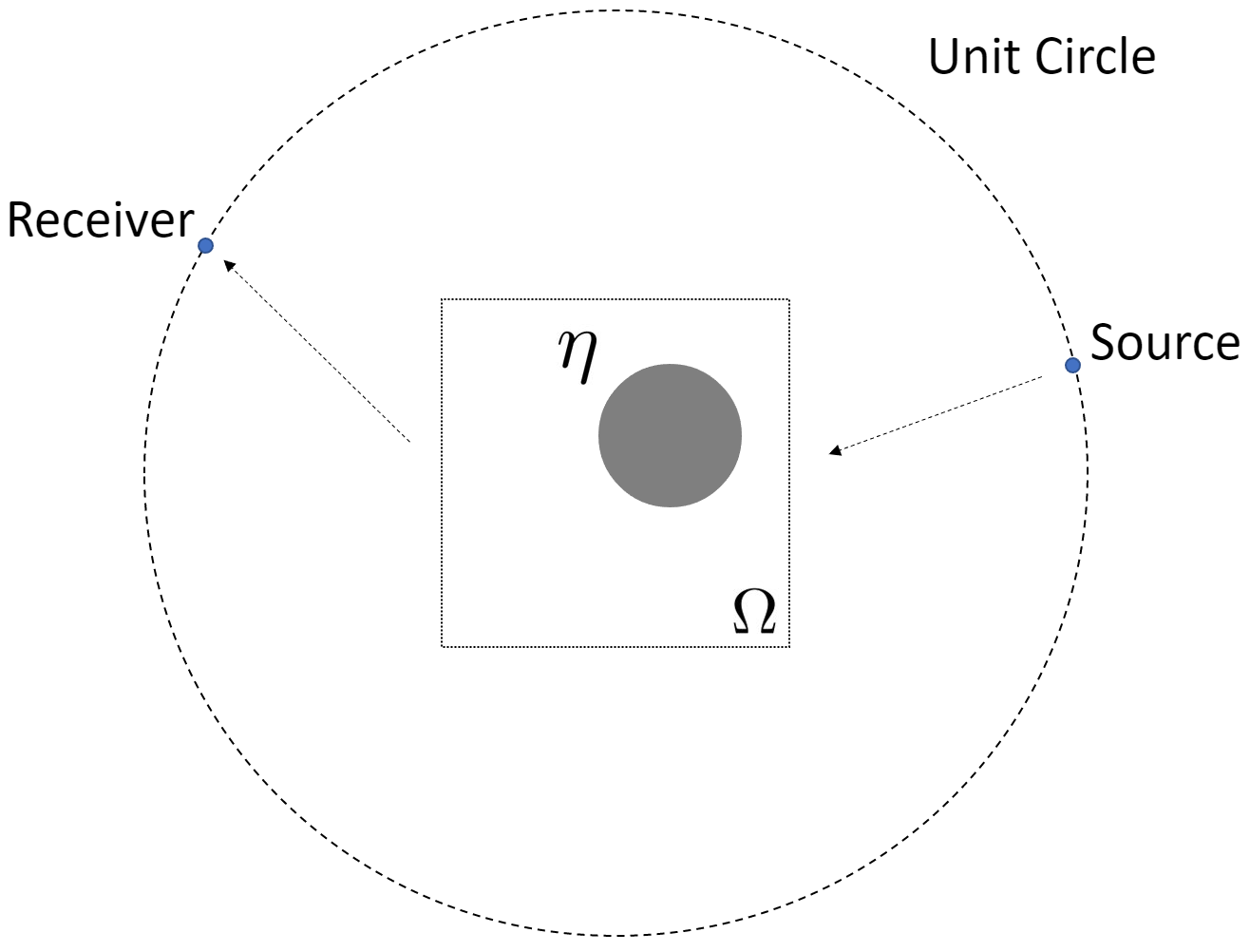}
\caption{Data generation process for the far-field pattern problem.}
\label{genp}
\end{figure}

\section{Preliminary of Inverse Scattering}
\label{sec2}
\subsection{Background}
\label{subsec2.1}

In this section, we discuss the forward model for the inverse scattering problem. The inhomogeneous media scattering problem at a fixed wavenumber $k$ is modeled by Equation \eqref{holm}. It is assumed that a scatterer $\eta(x)$ is compactly supported in a domain $\Omega$ (see Figure \ref{genp} for visualization). Typically, in a numerical solution of the Helmholtz operator, $\Omega$ is discretized by a Cartesian grid $X\subset\Omega$ at the rate of a few points per wavelength. Assume that $X$ has $N\times N$ grid points and $\{x\}_{x\in X}$ is used to denote the discretization points of $X$. 
After discretization, the scatterer field $\eta$ can be treated as a vector in $\mathbb{R}^{N^2}$ evaluated at $\{x\}_{x\in X}$.

\subsection{Far Field Pattern}
\label{subsec2.2}
In this work, we address the problem under the far-field assumption, a widely used approach in wave scattering problems. To simplify the domain, we rescale \( \Omega \) to fit within the unit circle, \( \mathbb{S}^1 \), where sensors are positioned along the circumference, with their locations specified by the corresponding angles. Let \( \sigma \in \mathbb{S}^1 \) represent a unit direction. A source in the direction \( \sigma \) emits an incoming plane wave described by \( e^{i\omega \sigma \cdot x} \), where \( \omega \) denotes the frequency. The relationship between frequency \( \omega \) and wavenumber \( k \) is:

\[
\omega = c k,
\]

where \( c \) is the wave velocity, assumed to be 1 for simplicity.

We define the scattered wave through the concept of an integral equation. Let \( u \) represent a new variable for the scattered field. The outgoing wave \( u_{\text{out}} \) can be expressed as a convolution integral with the Green's function:

\[
u_{\text{out}}(x) = \int_{\Omega} G(x,x') u(x') dx',
\]

where \( G(x,x') = \frac{i}{4} H_0^{(1)}(k|x - x'|) \) is the Green’s function for the 2D Helmholtz equation, and \( H_0^{(1)} \) is the Hankel function of the first kind, representing outgoing waves. This Green's function solution assumes the far-field approximation, where the wave is approximated as being planar in the asymptotic limit. The equation for the scattered wave becomes:

\[
(I - k^2 \eta G) * u = k^2 \eta u_{\text{in}},
\]

where the convolution operator \( * \) denotes integration over the domain \( \Omega \). In discretized form, the solution for \( u \) is given by:

\[
u = (I - k^2 \eta G)^{-1} k^2 \eta u_{\text{in}}.
\]

Once \( u \) is computed, the observed scattered wave \( u_{\text{out}} \) received by the sensors can be computed by applying the Green's function as:

\[
u_{\text{out}}(x) = \int_{\Omega} G(x, x') u(x') dx',
\]

where \( u_{\text{out}} \) represents the observed data \( d \).

For a given set of receivers \( \mathcal{R} \), source set \( \Sigma \), and frequency \( \omega \), the observed data \( d \) can be expressed as:

\begin{equation}
d = u_{\text{out}} = F_{\omega, \mathcal{R}, \Sigma}(\eta).
\label{gen}
\end{equation}

In the next section, based on the forward model \eqref{gen}, we introduce the proposed reinforcement learning scheme to solve the inverse problem.

\section{Reinforcement Learning Framework for Inverse Scattering}
\label{sec3}
In conventional inverse scattering data collection, sensors are typically positioned either randomly or uniformly around the unit circle, and wave frequencies are often selected through empirical methods. This paper introduces a reinforcement learning-based approach to develop a strategy that adaptively selects sensor angles and wave frequencies. The approach involves a sequential decision-making process where: 1) multiple sets of sensors and frequencies are selected step by step; 2) sensor placement and frequency selection are determined based on the reconstruction outcomes obtained from earlier data. This method aims to progressively enhance data collection, leading to improved reconstruction results. Furthermore, it employs tailored strategies for imaging different $\eta$ values, demonstrating its adaptability across various scenarios.

The remainder of this section is organized as follows. In Section \ref{subsec3.1}, we introduce the problem setting, laying the groundwork for the subsequent sections. Section \ref{subsec3.2} presents the Markov Decision Process (MDP) formulation of the problem, which allows us to apply reinforcement learning (RL) methods to address it. Section \ref{subsec3.3} reviews essential concepts and methods in RL and introduces the specific RL algorithm used to optimize the policy within the MDP described in Section \ref{subsec3.2}. In Section \ref{subsec3.4}, we describe the solver employed for scatterer reconstruction, which is a critical component of the RL algorithms discussed in Section \ref{subsec3.3}. Sections \ref{subsec3.5} and \ref{subsec3.6} detail the architecture of the policy network and the value network, respectively. Finally, Sections \ref{subsec3.7} and \ref{subsec3.8} explain the training and testing procedures of our RL model, integrating the content from Sections \ref{subsec3.1} through \ref{subsec3.6}.


\subsection{Problem Setting}
\label{subsec3.1}
Without loss of generality, we assume the true scatterer $\eta(x)$ is compactly supported within a unit square $\Omega$ centered at the origin, with all probes placed on the unit circle $\mathbb{S}^1$ (encompassing $\Omega$). In the reinforcement learning framework described later, an action $a_t$ determines the sensor locations and frequency choices at time $t$. We uniformly discretize the unit circle $\mathbb{S}^1$ and define $\sigma^{a_t} \in \mathbb{R}^{360}$ as a one-hot vector specifying the sensor location. $\sigma^{a_t}$ has a single non-zero entry indicating the sensor's angle on $\mathbb{S}^1$. In our experiment, one sensor is placed on the unit circle at each step until $T$ sensors are positioned, with $T$ being a user-defined number. At each step, sensors emit incident waves and receive scattered waves. At time $t$, with sensors sequentially placed at $\sigma^{a_i}$ $(i=1,2,\ldots,t-1)$, a new sensor $\sigma^{a_t}$ is added to the unit circle. A wave field of frequency $\omega^{a_t}$ is launched from this new sensor and propagates through the domain $\Omega$. Sensors at $\sigma^{a_1}, \ldots, \sigma^{a_t}$ then receive the scattered wave. Each sensor functions as both a source and a receiver. The receiver set at time $t$ is denoted as $\mathcal{R}^{a_{t}}$, and the source set at time $t$ is $\Sigma^{a_{t}}$. Thus, $\mathcal{R}^{a_{t}} = \{\sigma^{a_1}, \ldots, \sigma^{a_t}\}$ and $\Sigma^{a_{t}} = \{\sigma^{a_t}\}$. We approximate the observed data at time $t$ using \eqref{gen}: $d_{t} = F_{\omega^{a_{t-1}}, \mathcal{R}^{a_{t-1}}, \Sigma^{a_{t-1}}}(\eta)$. Notably, the frequency $\omega^{a_{t}}$ at step $t$ can be different from the frequency at other steps. After $T$ steps, the data collection process concludes, and we reconstruct the scatterer $\eta$ using all recorded measurements. The entire data collection procedure, comprising $T$ sequential steps, is termed an episode. The objective of this paper is to enhance reconstruction quality while limiting the number of probes used. Our approach involves learning to improve a given data collection strategy.

The original problem of determining sensor location and wave frequency is a combinatorial optimization problem and is NP-hard. We'll formulate the problem as an MDP in Section \ref{subsec3.2}, which can be further solved by reinforcement learning methods.

\subsection{Markov Decision Process Formulation}
\label{subsec3.2}
The procedure of deciding sensor locations and frequency values in inverse scattering is a sequential decision problem, where one needs to make a choice of angle and frequency at each step. Thus it can be formulated as a Markov Decision Process (MDP). In this way, we can use RL to solve the problem efficiently. We now elaborate on how to formulate our problem as an MDP:
\begin{enumerate}
\item \textbf{State} at time $t$ is $s_t=(I_1,I_2,\ldots,I_t)$, where $I_t=(d_t,u_{t},T+1-t)$ represents the information at step $t$. The first term $d_t=F_{\omega^{a_{t-1}} \mathcal{R}^{a_{t-1}},\Sigma^{a_{t-1}}}(\eta)\in\mathbb{C}^{1\times (t-1)}$ is the collected observation data at step $t$. The size $1\times(t-1)$ comes from the fact that at time $t$, we send out waves from a single source to the rest $t-1$ receivers. Measurements up to time $t$ are all included in state $s_t$ because the reconstruction at step $t$ relies on all the previously collected data. The second term $u_{t}\in\mathbb{R}^{360}$ is a vector recording all the angles where sensors have already been placed by time $t$ and the corresponding wave frequencies. The $j$-th entry of $u_t$ denotes the angle $\frac{2j\pi}{360}$. If the $j$-th angle is selected at any step $k$ $(k\le t)$, then the $j$-th entry of $u_{t}$ is $\omega^{a_{k}}$, denoting the frequency selected at the $k$-th step. If no wave is sent from a specific angle, the entry in $u_t$ corresponds to that angle is $0$. The last term $T+1-t$ means the number of sensors left to be placed. 
\item \textbf{Action} at time $t$, denoted as $a_t$, defines the choices for angles and frequencies $\sigma^{a_t}$ and $\omega^{a_t}$. Here, $\sigma^{a_t} \in \mathbb{R}^{360}$ is a one-hot vector indicating the angle of the new sensor on the unit circle at time $t$, and $\omega^{a_t} \in \mathbb{R}$ represents the frequency of the incident wave. Specifically, $u_t$ is defined in terms of $\sigma^{a_t}$ and $\omega^{a_t}$ as $u_t = \sum_{j=1}^{t-1} \omega^{a_j} \sigma^{a_j}$.
\item \textbf{Transition model}. The state $s_t$ and action $a_t$ allow us to compute the deterministic state $s_{t+1}$ in a noise-free model. According to \eqref{gen}, the new measurement $d_{t+1} = F_{\omega^{a_{t}}, \mathcal{R}^{a_{t}}, \Sigma^{a_{t}}}(\eta)$ can be calculated given $s_t$ and $a_t$. Concurrently, $u_{t+1} = u_{t} + \omega^{a_{t}} \sigma^{a_t}$. Thus, $I_{t+1} = (d_{t+1}, u_{t+1}, T-t)$ and the new state $s_{t+1} = (I_1, \ldots, I_{t+1})$.
\item \textbf{Reward} at time $t$, denoted as $r_t$, is defined as the increment in the Peak Signal to Noise Ratio (PSNR) value of the reconstruction compared to the previous step. PSNR is a common metric for assessing image reconstruction quality. Here, we use its increment to quantify the improvement due to the new action. Let $\hat{\eta}_{t}$ be the reconstruction at step $t$ and $\eta$ be the true scatterer, then $r_t = \text{PSNR}(\hat{\eta}_{t}, \eta) - \text{PSNR}(\hat{\eta}_{t-1}, \eta)$. In this paper, the scatterer is reconstructed using a regularization-based optimization method, which will be introduced later. We note that better results might be achieved with more sophisticated reconstruction methods.
\end{enumerate}


\subsection{Reinforcement Learning Algorithm}
\label{subsec3.3}
In RL, an agent interacts with the environment to obtain a sequence of data, based on which the agent learns a policy to maximize a certain accumulated reward to finish a task. Given an interaction trajectory between agent and environment $\tau=(s_1, a_1, r_1, s_2, a_2, r_2,\ldots)$, the total reward over time is
\begin{align*}
G(\tau) = r_1+\gamma r_2 + \gamma^2 r_3 + \ldots,
\end{align*}
where $\gamma\in [0,1]$ is a discount rate. We use $\gamma=1$ since the reward at each step is of equal importance in our application. The policy in the MDP is defined as the conditional probability $\pi(a|s)$ for $a_t=a$ and $s_t=s$. $\pi(a|s)$ denotes the probability of taking action $a$ at step $t$ while the state is $s$. The value function of a state $s$ following a certain policy $\pi$ is given as
\begin{align*}
v_{\pi}(s) = \mathbb{E}_{\tau}[r_{1}+\gamma r_{2}+\gamma^2 r_{3}+\ldots|s_1=s,\tau\sim\pi].
\end{align*}
Similarly, the value function of a state-action pair $(s,a)$ is defined as
\begin{align*}
q_{\pi}(s,a)=\mathbb{E}_{\tau}[r_{1}+\gamma r_{2}+\gamma^2 r_{3}+\ldots|s_1=s,a_1=a,\tau\sim\pi].
\end{align*}
In an MDP, the RL algorithm aims to find an optimal policy that maximizes the expected value of initial state $s_1$ following a probability distribution $p(s)$:
\begin{align*}
\max\limits_{\pi}\mathbb{E}_{s_1\sim p(s)}[v_{\pi}(s_1)].
\end{align*}
Currently, model-free RL algorithms mainly fall into two categories: value-based methods and policy-based methods. Value-based algorithms learn the state or state-action value and act by choosing the best action in the state, which requires comprehensive exploration. For example, Q-learning \citep{qlearning} learns the optimal Q function through the Bellman Equation and chooses a greedy action, which maximizes the learned Q function. Since the maximization requires searching on the action space, it will become slow and imprecise if the actions are continuous. This means that value-based algorithms are more suitable for discrete actions. On the other hand, policy gradient methods \citep{policy1,policy2} are more suitable for continuous actions because they directly optimize a parameterized policy under a surrogate objective. Though the variables (angles and frequencies) in our current setting are discrete, we adopt the policy-based method for possible future extensions to continuous cases.

Specifically, we use policy gradient methods that directly optimize a policy $\pi_{\theta}$, parameterized by a neural network, under a certain objective function. Policy gradient methods seek to maximize the performance of $\pi_{\theta}$ via stochastic updates, whose expectation approximates the gradient of the performance measure with respect to $\theta$. In our experiments, we use the Proximal Policy Optimization (PPO) algorithm \citep{ppo}. Given an old policy $\pi_{\theta_{\text{old}}}$ and a new policy $\pi_{\theta}$, let $\gamma_{\theta}$ denote the probability ratio $\gamma_{\theta}(s,a)=\frac{\pi_{\theta}\left(a \mid s\right)}{\pi_{\theta_{\text {old }}}\left(a \mid s\right)}$. An advantage function of policy $\pi_\theta$ is introduced as 
\begin{equation}\label{eqn:af}
\alpha_{\pi_\theta}(s,a)=q_{\pi_\theta}(s,a)-v_{\pi_\theta}(s).
\end{equation}
The objective function to optimize is then defined as:
\begin{align*}
L^{\operatorname{clip}}(\theta)=\mathbb{E}_{s,a\sim\pi_{\theta_{\text{old}}}}\left[\min \left(\gamma_{\theta}(s,a)\alpha_{\pi_{\theta_{\text{old}}}}(s,a), \operatorname{clip}\left(\gamma_{\theta}(s,a), 1-\epsilon, 1+\epsilon\right) \alpha_{\pi_{\theta_{\text{old}}}}(s,a)\right)\right],
\end{align*}
where $\epsilon$ is a hyperparameter and $\text{clip}(x,y,z)=\min(\max(x,y),z)$. This method employs clipping to avoid destructively large policy updates, which retains 
the stability and reliability of trust-region methods but is much simpler to implement in practice.

At the end of each step, where the model determines a position and frequency, we gather new observations from the sensor, which are then used to update the scatterer reconstruction. Based on this updated reconstruction, the reward is calculated, allowing the model to be further refined in the subsequent time step.

\subsection{Reconstruction}
\label{subsec3.4}
According to \eqref{gen}, for a choice of frequency $\omega^{a_{t}}$, receiver set $\mathcal{R}^{a_{t}}$ and source set $\Sigma^{a_{t}}$ at step $t$, the approximate measurement $d_{t+1}$ received by the receivers is given by
\begin{align*}
d_{t+1}= F_{\omega^{a_{t}},\mathcal{R}^{a_{t}},\Sigma^{a_{t}}}(\eta),
\end{align*}
where $\eta$ is the true scatterer. We reconstruct $\eta(x)$ by minimizing an $\ell_2$ loss of data discrepancy with an $\ell_1$ penalization to encourage sparsity:
\begin{align*}
\min_{\hat{\eta}}\ \|d-F_{\omega,\mathcal{R},\Sigma}(\hat{\eta})\|_{2}^2 + \lambda \|\hat{\eta}\|_{1},
\end{align*}
where $\lambda>0$ is a hyperparameter. Due to the small number of measurements, we observe that the use of a regularization like $\ell_1$ significantly improves the results. A new reconstruction is generated at each step in order to compute a reward of the RL model. Given state $s_t$ and action $a_t$, the reconstruction $\hat{\eta}_{t}$ obtained at time $t$ is:
\begin{align}
\begin{aligned}
\hat{\eta}_{t}&=\underset{\hat{\eta}}{\text{argmin}}\ \sum\limits_{i=1}^{t}\|d_{i+1}-F_{\omega^{a_{i}},\mathcal{R}^{a_{i}},\Sigma^{a_{i}}}(\hat{\eta})\|_{2}^2 + \lambda \|\hat{\eta}\|_{1}, \label{rec}
\end{aligned}
\end{align}
which takes advantage of all the previously collected measurements.

In the numerical experiment, we set $\lambda=1 \mathrm{e}{-5}$ and use the L-BFGS algorithm to solve the optimization problem in \eqref{rec} for the reconstruction at each step. Starting from an initial estimate \(\hat{\eta}\), we stop the L-BFGS algorithm after $3$ iterations at each step. It is crucial to use a warm start to achieve good results. We use the reconstruction result from the previous step as the initialization for the current step. This approach significantly reduces the computational cost and ensures convergence. For the warm-start algorithm, we simply use zero as the initial guess. 

Besides, L-BFGS can be time-consuming, and using the Adam optimizer can greatly reduce computational time, while slightly compromising the reconstruction quality. 

In the previous sections, we formulate the original problem as an MDP and introduce how to compute the terms in the MDP. In order to apply policy gradient methods, we need to build a policy network that parameterizes the policy to learn.

\subsection{Policy Network}
\label{subsec3.5}
\begin{figure}[H]
    \centering
    \includegraphics[scale=0.2]{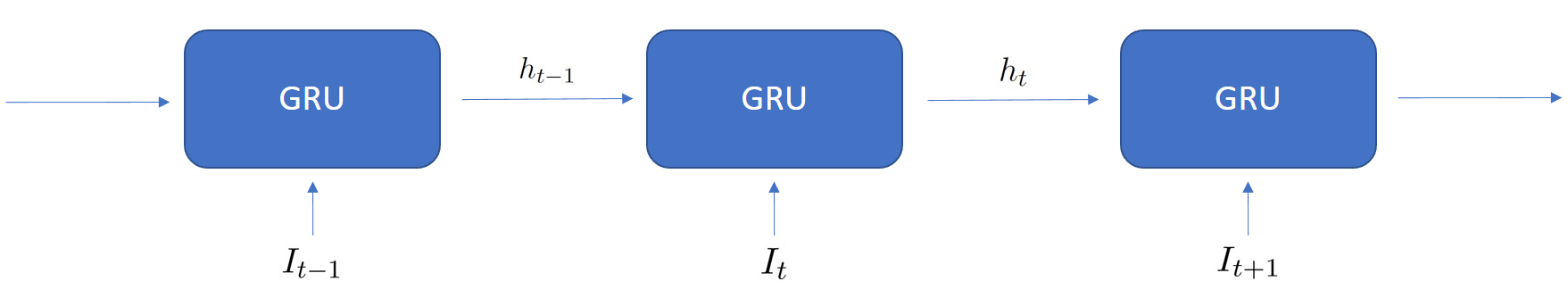}
    \caption{Structure of Recurrent Neural Network. Each GRU represents one layer, $I_t$ is the input of layer $t$. Each layer outputs a hidden state $h_t$ which is also the input of the next layer.}
    \label{rnns}
\end{figure}

To manage the increasing dimensionality of the state \(s_t\) over time, we use a Recurrent Neural Network (RNN) to parameterize the policy. This RNN structure allows us to store all past information in a hidden state while incorporating new information at each step. Specifically, we employ multi-layer Gated Recurrent Units (GRUs), as introduced by \cite{gru} and applied in \cite{ct}. GRUs are similar to Long Short-Term Memory (LSTM) units but with fewer parameters. The structure of the GRU is shown in Figure \ref{rnns}.

The policy network, denoted by \(\pi_{\theta_p}\), is parameterized by \(\theta_p\). The hidden state at layer \(t\) is represented by \(h_t^p\). Initially, a multi-layer perceptron (MLP) extracts features from the input \(I_t = (d_t, u_t, T + 1 - t)\). These features, along with the hidden state from the previous layer (\(h_{t-1}^p\)), are processed by the GRU.

To determine the sensor angle, the GRU output is further processed by another MLP. This MLP learns the policy for the angle based on the state information \(s_t\). Using a softmax function, its output is transformed into a categorical distribution of angles (a $360$-dimensional vector), where each entry represents the probability of placing a sensor at the corresponding angle. To ensure that each angle is chosen only once per episode, a mask is used to eliminate previously selected angles. The angle distribution is denoted by \(\pi_{\theta_p}^{\sigma}\), representing the policy for angle \(\sigma\).

During training, an angle is sampled from the distribution \(\pi_{\theta_p}^{\sigma}\), which is encoded as a one-hot 360-dimensional vector, with \(1\) indicating the chosen angle and \(0\) otherwise. The GRU output \(h_t^p\) and the selected angle \(\sigma^{a_t}\) are then combined into a single vector, which is used to learn the frequency policy. The choice of the wave frequency depends on both the current state \(s_t\) and the chosen angle \(\sigma^{a_t}\). In our experiment, the frequency is selected from four possible values. Thus, the combined vector is processed by another MLP, which outputs a 4-dimensional vector representing the categorical distribution of frequencies. The distribution is denoted by \(\pi_{\theta_p}^{\omega|\sigma}\), indicating that the frequency policy \(\omega\) depends on the angle \(\sigma\).

\begin{figure}
    \centering
    \includegraphics[scale=0.2]{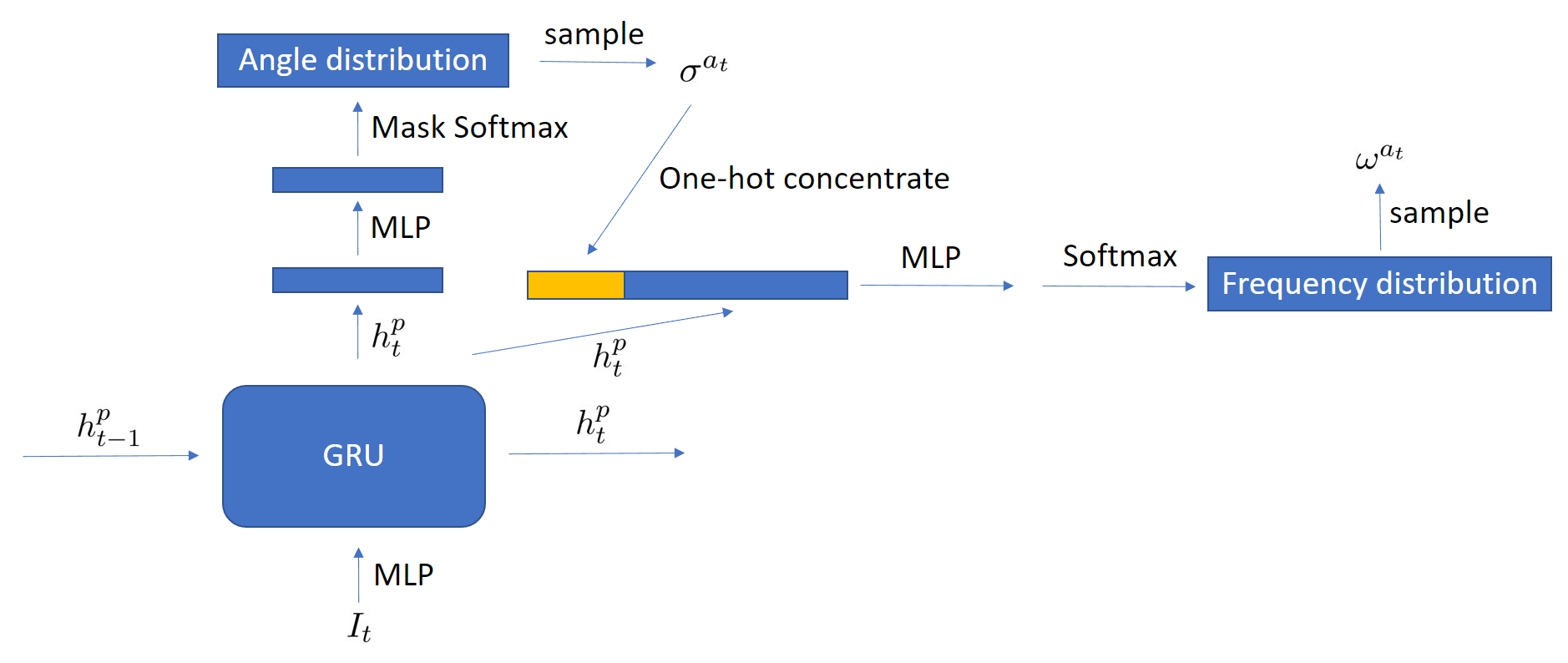}
    \caption{Structure of policy network $\pi_{\theta_p}$. $h_t^{p}$ represents the hidden state of the policy net, which is the output of GRU at layer $t$. $I_t=(d_t,u_t,T+1-t)$. We use $h_t^p$ as the input of another perceptron and generate a 360-dim categorical distribution of angle through the softmax function with a mask removing angles that have been chosen. This distribution is the angle policy $\pi_{\theta_p}^{\sigma}$. Then we randomly generate an angle $\sigma^{a_t}$ based on distribution and combine its one-hot concentrate with $h_t^{p}$ as the input of another MLP, which gives rise to another categorical distribution of frequency. This distribution represents the frequency policy given angle $\pi_{\theta_p}^{\omega|\sigma}$. Finally, we use this to randomly generate a frequency $\omega^{a_t}$.}
    \label{policy}
\end{figure}

Finally, a frequency is randomly sampled from the distribution \(\pi_{\theta_p}^{\omega|\sigma}\). The structure of the policy network is illustrated in Figure \ref{policy}. In summary, the policy network learns the policy \(\pi_{\theta_p}(a|s)\) for an action given a state, which consists of an angle policy \(\pi_{\theta_p}^{\sigma}\) and a frequency policy \(\pi_{\theta_p}^{\omega|\sigma}\). During training, the angles and frequencies generated by the policy networks are random samples from these two distributions.

\subsection{Value Network}
\label{subsec3.6}
\begin{figure}
    \centering
    \includegraphics[scale=0.2]{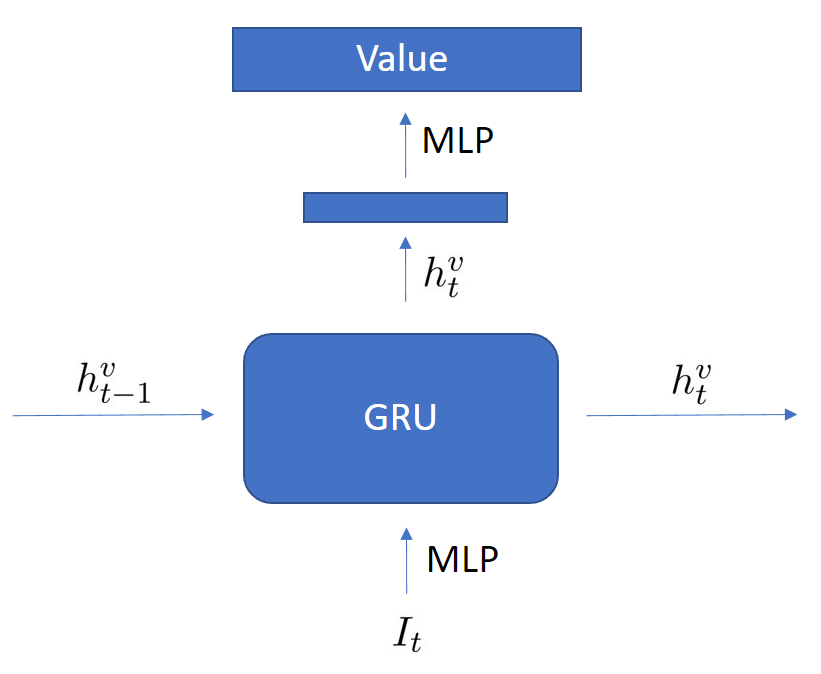}
    \caption{Structure of value network $\gamma_{\theta_v}$. $h_t^{v}$ represents the hidden state of value net, which is the output of GRU at layer $t$. $I_t=(d_t,u_t,T+1-t)$. We use $h_t^{v}$ as the input of another perceptron and generate $\hat{v}_{\pi_{\theta_p}}(s_t)$, which is the estimate of the value of current state $s_t$ under the policy $\pi_{\theta_p}$ parameterized by policy network. }
    \label{value}
\end{figure}

In addition to the policy network, we also build a value network \(\gamma_{\theta_v}\) parameterized by trainable parameters \(\theta_v\), with a similar structure to the policy network, to approximate the value function of states. The value function approximation is required for evaluating the advantage function of the policy \(\pi_\theta\) in \eqref{eqn:af}. The design of the value network is visualized in Figure \ref{value}. We use \(h_t^v\) to represent the output of the value network after the \(t\)-th layer. At the \(t\)-th step, an MLP is used to extract features from the current input \(I_t = (d_t, u_t, T + 1 - t)\). These features, along with the output of the GRU from the previous step (the hidden state \(h_{t-1}^v\)), are processed by a GRU to generate \(h_t^v\).

The difference between the policy network and the value network lies in how they process \(h_t^v\) to generate useful information. In the value network, the GRU output \(h_t^v\) is processed by an MLP to generate a deterministic estimate of the value function \(v_{\pi_{\theta_p}}(s)\), rather than a distribution as in the policy network. The estimated value of a state given by the value network is denoted by \(\hat{v}_{\pi_{\theta_p}}(s)\).

With the policy network and the value network, we can train the RL model using policy gradient methods.

\subsection{Training Procedure}
\label{subsec3.7}
We now present the training procedure for the policy network \(\pi_{\theta_p}\) and the value network \(\gamma_{\theta_v}\). The training set comprises randomly generated scatterers. A scatterer \(\eta\) is randomly selected from this set to generate a interaction trajectory \(\tau = (s_1, a_1, r_1, \ldots, s_T, a_T, r_T)\).

At the start of each episode, the state is initialized with \(d_1 = 0\), \(u_1 = 0\), and \(h_1^p = h_1^v = 0\), resulting in an initial state of \(s_1 = I_1 = (d_1, u_1, T)\). At each step \(t\), given the hidden state \(h_{t-1}^p\) from the previous step and the current information \(I_t\), the policy network learns a policy \(\pi_{\theta_p}(a_t|s_t)\). Concurrently, the value network estimates the value of state \(s_t\) under the current policy, denoted as \(\hat{v}_{\pi_{\theta_p}}(s_t)\).

During training, angles and frequencies are randomly sampled based on the policy \(\pi_{\theta_p}(a_t|s_t)\), promoting thorough exploration of the action space for faster convergence. A new sensor is placed at angle \(\sigma^{a_t}\) to launch an incident wave of frequency \(\omega^{a_t}\). The receiver set at step \(t\) is updated \(\mathcal{R}^{a_t} = \{\sigma^{a_1}, \ldots, \sigma^{a_t}\}\), and the sensor set is \(\Sigma^{a_t} = \{\sigma^{a_t}\}\).

After obtaining a new measurement \(d_{t+1}\), we compute \(u_{t+1} = u_t + \omega^{a_t}\sigma^{a_t}\) and update the state to \(s_{t+1} = (d_{t+1}, u_{t+1}, T - t)\). The reconstruction \(\hat{\eta}_t\) at time \(t\) is derived from all previously collected measurements. The reward \(r_t\) is computed as the increment in PSNR of the reconstruction: \(r_t = \text{PSNR}(\hat{\eta}_t, \eta) - \text{PSNR}(\hat{\eta}_{t-1}, \eta)\).

With \(r_t\) and \(s_{t+1}\) available, the value network estimates \(\hat{v}_{\pi_{\theta_p}}(s_{t+1})\) to compute the advantage function \(\alpha_{\pi_{\theta_p}}(s_t, a_t) \approx \hat{v}_{\pi_{\theta_p}}(s_{t+1}) + r_t - \hat{v}_{\pi_{\theta_p}}(s_t)\), denoted as \(\hat{\alpha}_{\pi_{\theta_p}}(s_t, a_t)\). These estimates are then utilized in the Proximal Policy Optimization (PPO) algorithm.

The episode concludes after \(T\) steps, resulting in a trajectory \(\tau = (s_1, a_1, r_1, \ldots, s_T, a_T, r_T)\). We generate multiple episodes in parallel and train the networks using mini-batches of these episodes. From these simulations, we compute the PPO surrogate objective and optimize our networks the Adam optimizer. Hyperparameters and specific settings are detailed in Section \ref{sec4}. PPO's advantage lies in its capacity for multiple optimization steps using a few trajectories without inducing large policy updates. The detailed algorithm is provided in Algorithm \ref{train_procedure}.

Following training, we evaluate the model's performance on a new test set, adhering to a slightly modified procedure from training to assess the model's generalization ability.

\subsection{Test Procedure}
\label{subsec3.8}
Given the fully trained policy network \(\pi_{\theta_p}\), we proceed to evaluate the RL model on a set of scatterers that are similar to those in the training set. During the testing phase, a scatterer \(\eta\) is randomly selected from the test set, and the policy network generates an interaction trajectory \(\tau = (s_1, a_1, r_1, \ldots, s_T, a_T, r_T)\). The initialization parameters remain consistent with the training phase: \(d_1 = 0\), \(u_1 = 0\), \(h_0^p = 0\), and \(\hat{\eta}_0 = 0\).

At each step \(t\), the policy network determines the policies for selecting angles \(\pi_{\theta_p}^{\sigma}\) and frequencies \(\pi_{\theta_p}^{\omega}\) based on the hidden state \(h_t^p\) and the current information \(I_t = (d_t, u_t, T + 1 - t)\). These policies provide probability distributions for angles and frequencies. It is important to note that during RL training, angles and frequencies are randomly chosen according to their probability distributions to encourage exploration of the action space. However, during RL model testing, we select the angle and frequency with the highest probability to optimize reconstruction quality.

Once an action is selected, a new sensor is placed, and the state is updated: \(d_{t+1} = F_{\omega^{a_t}, \mathcal{R}^{a_t}, \Sigma^{a_t}}(\eta)\) and \(u_{t+1} = u_t + \omega^{a_t}\sigma^{a_t}\). This process is repeated for \(T\) steps, culminating in a final reconstruction.

The testing phase not only validates the effectiveness of the trained policy network but also ensures that the model generalizes well to new scatterers. By consistently selecting the highest probability actions during testing, we aim to achieve the most accurate and high-resolution reconstructions, demonstrating the robustness and reliability of the RL model.

\begin{algorithm}
\caption{The Training Procedure of Our Reinforcement Learning Algorithm}
\label{train_procedure}
\begin{algorithmic}[1]
\STATE {\bf Require}: Training sample size \(k\), training samples \(\{\eta_i\}_{1 \le i \le k}\), grid size \(N\), total number of sensors \(T\), policy network \(\pi_{\theta_p}\), value network \(\gamma_{\theta_v}\), clipping constant \(\epsilon\)
\STATE {\bf For} epoch \(= 0, 1, 2, \ldots\)
\STATE \qquad Initialization: \(d_1 = 0\), \(u_1 = 0\), \(s_1 = I_1 = (d_1, u_1, T)\), \(h_0^p = 0\), \(h_0^v = 0\), \(\hat{\eta}_0 = 0\), \(\hat{v}_{\pi_{\theta_p}}(s_1) = 0\), randomly choose a sample \(\eta\) from \(\{\eta_i\}_{1 \le i \le k}\)
\STATE \qquad {\bf For} \(t = 1, \ldots, T\):
\STATE \qquad \qquad Given \(h_{t-1}^p\) and \(I_t\), use policy networks to generate policies \(\pi_{\theta_p}^{\sigma}\) and \(\pi_{\theta_p}^{\omega|\sigma}\), then update \(h_t^p\)
\STATE \qquad \qquad Randomly sample an angle \(\sigma^{a_t}\) from \(\pi_{\theta_p}^{\sigma}\) and a frequency \(\omega^{a_t}\) from \(\pi_{\theta_p}^{\omega|\sigma}\)
\STATE \qquad \qquad Update the receiver set \(\mathcal{R}^{a_t} = \{\sigma^{a_1}, \ldots, \sigma^{a_t}\}\) and the source set \(\Sigma^{a_t} = \{\sigma^{a_t}\}\)
\STATE \qquad \qquad Update the state \(s_{t+1}\)
\STATE \qquad \qquad Given \(h_{t-1}^v\) and \(I_t\), use the value network to approximate \(\hat{v}_{\theta_p}(s_{t+1})\), then update \(h_t^v\)
\STATE \qquad \qquad Reconstruct the scatterer \(\hat{\eta}_t\)
\STATE \qquad \qquad Compute the reward \(r_t = \text{PSNR}(\hat{\eta}_t, \eta) - \text{PSNR}(\hat{\eta}_{t-1}, \eta)\)
\STATE \qquad \qquad Approximate the advantage function \(\hat{\alpha}_{\pi_{\theta_{p\text{old}}}}(s_t, a_t) = \hat{v}_{\pi_{\theta_p}}(s_{t+1}) + r_t - \hat{v}_{\pi_{\theta_p}}(s_t)\)
\STATE \qquad {\bf End for}
\STATE \qquad Compute \(\gamma_{\theta_p}(s_t, a_t) = \frac{p^{\sigma^{a_t}}_{\theta_p} p^{\omega^{a_t}}_{\theta_p}}{p^{\sigma^{a_t}}_{\theta_\text{old}} p^{\omega^{a_t}}_{\theta_\text{old}}}\)
\STATE \qquad Evaluate \(L^{\operatorname{clip}}(\theta_p) = \frac{1}{T} \sum_{t=1}^{T} \min(\gamma_{\theta_p}(s_t, a_t) \hat{\alpha}_{\pi_{\theta_{p\text{old}}}}(s_t, a_t), \text{clip}(\gamma_{\theta_p}(s_t, a_t), 1 - \epsilon, 1 + \epsilon) \hat{\alpha}_{\pi_{\theta_{p\text{old}}}}(s_t, a_t))\)
\STATE \qquad Evaluate \(L(\theta_v) = \frac{1}{T} \sum_{t=1}^{T} (\hat{\alpha}_{\pi_{\theta_{p\text{old}}}}(s_t, a_t) + \hat{v}_{\pi_{\theta_p}}(s_t) - \hat{v}_{\theta_v}(I_t))^2\)
\STATE \qquad Optimize \(\theta_p\) and \(\theta_v\) for \(L^{\operatorname{clip}}(\theta_p)\) and \(L(\theta_v)\)
\STATE {\bf End for}
\STATE {\bf Return}: Trained policies \(\pi_{\theta_p}\) and \(\gamma_{\theta_v}\)
\end{algorithmic}
\end{algorithm}

\section{Numerical Experiments}
\label{sec4}
In this section, we will use several experiments to show that our model is capable of improving a given strategy through reinforcement learning. We focus on the far-field pattern and use our model to improve a couple of strategies. 

\subsection{Setting}
\label{subsec4.1}
In the far-field pattern analysis, we define the domain as \(\Omega = [-0.5, 0.5] \times [-0.5, 0.5]\), with the scatterer field \(\eta(x)\) discretized on a grid of \(s \times s\) points, where \(s = 32\) and \(s = 64\). In each experiment, we generate 6000 different scatterers, using 5000 for training and the remaining 1000 for testing. Sensor positions are specified with integer angles in the range \([0, 360)\), and the possible wavenumbers \(k\) are \(s/4, s/2, 3s/4,\) and \(s\). The corresponding wavelengths are \(\frac{8\pi}{s}, \frac{4\pi}{s}, \frac{8\pi}{3s},\) and \(\frac{2\pi}{s}\), respectively. Our algorithm allows for more frequency choices to potentially improve resolution, but we limit the options here for computational efficiency. We deploy a total of \(T = 10\) probes for sensing, with each probe placed individually. The value of \(T\) is a user-defined hyperparameter that depends on the specific requirements for reconstruction resolution. A larger \(T\) generally yields better reconstruction results.

In our experimental setup, we assume the measurements approximately follow the model outlined in Equation \eqref{gen}. When a new sensor is introduced and new measurements are collected, we use the L-BFGS algorithm to optimize the objective function described in \ref{rec}. Only three iterations of L-BFGS are run, and the resulting output serves both as the reconstruction for computing the reward at the current step and as the initialization for optimization in the subsequent step. The penalization constant \(\lambda\), introduced in Section \ref{subsec3.3}, is set to 0.1.

In the policy network, a multi-layer GRU with 3 recurrent layers is employed, with each layer containing 256 neurons. The angle MLP comprises 1 hidden layer with 512 neurons, while the frequency MLP consists of 2 hidden layers, each with 512 neurons. In the value network, the value MLP features 1 hidden layer of 512 neurons. Additionally, the MLPs in both the policy and value networks that extract features from \(s_t\) include 2 hidden layers, each containing 512 neurons. We use the Adam optimizer \citep{kingma2014adam} to optimize the parameters of the policy and value networks, with a learning rate of 0.0004. The coefficients used for computing running averages of the gradient and its square are \(\beta_1=0.9\) and \(\beta_2=0.999\). The policy and value networks are trained with PPO in each experiment for 1500 iterations. 


We utilize three commonly employed initial strategies. The first strategy uniformly selects angles \(0, \frac{\pi}{5}, \cdots, \frac{9\pi}{5}\) with a fixed frequency. The second strategy involves randomly selecting angles while maintaining a fixed frequency. The third strategy selects specific angles \(0, \frac{\pi}{3}, \frac{4\pi}{9}, \frac{5\pi}{9}, \frac{2\pi}{3}, \pi, \frac{4\pi}{3}, \frac{13\pi}{9}, \frac{14\pi}{9}, \frac{5\pi}{3}\), using a lower frequency at angles \(0\) and \(\pi\), and higher frequencies for the remaining angles. We first pretrain the model to learn these strategies, and subsequently train the model using our RL framework. This approach aims to demonstrate that RL can enhance any given strategy.

A fixed wavenumber \(k\) is used throughout an entire episode. We employ cross-validation to select the wavenumber that results in the lowest error for each method. To quantify reconstruction accuracy, we use the Mean Squared Error (MSE):

\[
\text{MSE} = \frac{1}{N^2}\sum_{x \in X}(\eta(x) - \hat{\eta}(x))^2,
\]

and the Peak Signal-to-Noise Ratio (PSNR):

\[
\text{PSNR} = 20 \log_{10}\left(\frac{\text{MAX}_{f}}{\sqrt{\text{MSE}}}\right),
\]

where \(\text{MAX}_{f}\) is the maximum possible pixel value of an image. A method is deemed satisfactory if it produces a small MSE or a large PSNR.

\begin{figure}
\centering
\setcounter{subfigure}{0}
\subfigure[Scatterer $1$ Groundtruth]{\label{fig:a}\includegraphics[width=45mm]{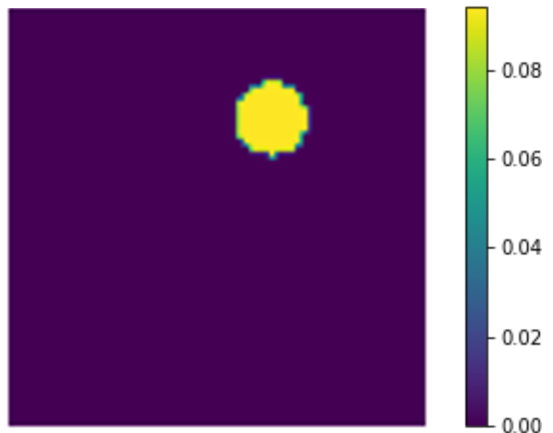}}
\hfill/\
\subfigure[Reconstruction by the pretrained strategy (PSNR$=147.47$)]{\label{fig:b}\includegraphics[width=45mm]{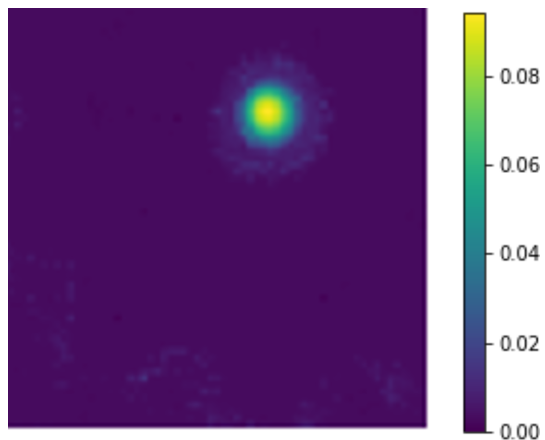}}
\hfill
\subfigure[reconstruction by RL-trained strategy (PSNR$=150.55$)]{\label{fig:c}\includegraphics[width=45mm]{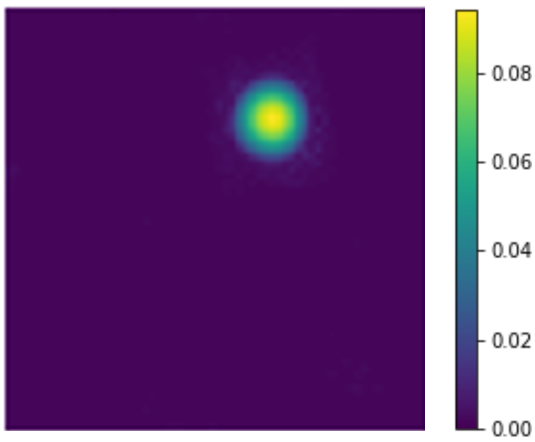}}
\subfigure[Scatterer $2$ Groundtruth]{\label{fig:d}\includegraphics[width=45mm]{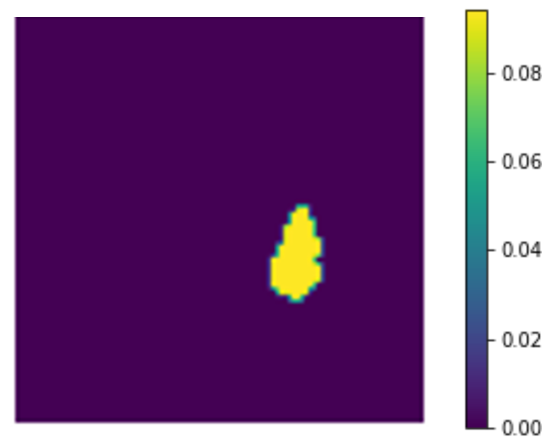}}
\hfill
\subfigure[Reconstruction by the pretrained strategy (PSNR$=108.96$)]{\label{fig:e}\includegraphics[width=45mm]{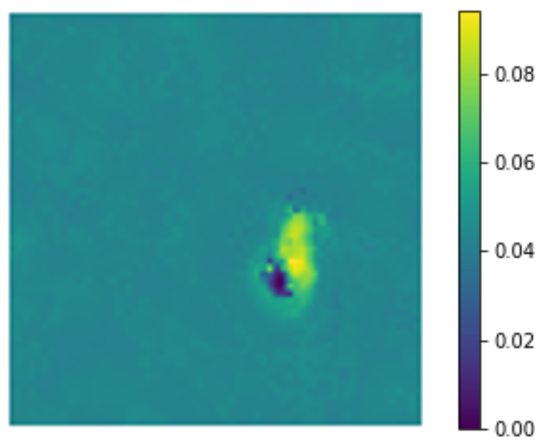}}
\hfill
\subfigure[reconstruction by RL-trained strategy (PSNR$=113.54$)]{\label{fig:f}\includegraphics[width=46mm]{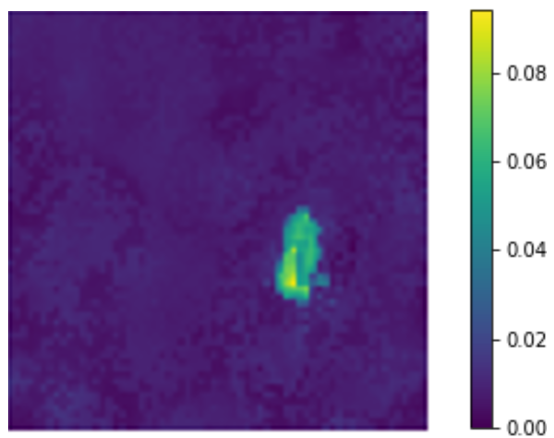}}
\subfigure[Scatterer $3$ Groundtruth]{\label{fig:g}\includegraphics[width=45mm]{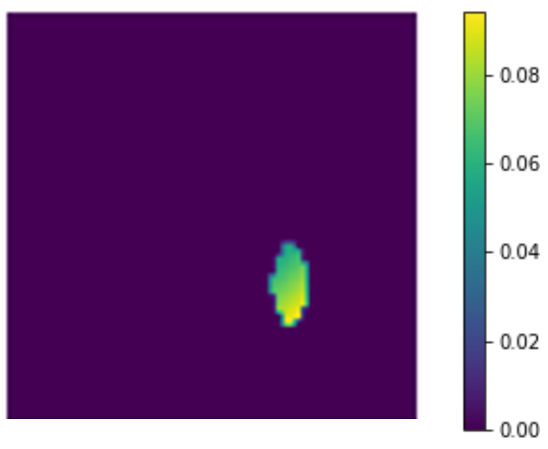}}
\hfill
\subfigure[Reconstruction by the pretrained strategy (PSNR$=108.98$)]{\label{fig:h}\includegraphics[width=45mm]{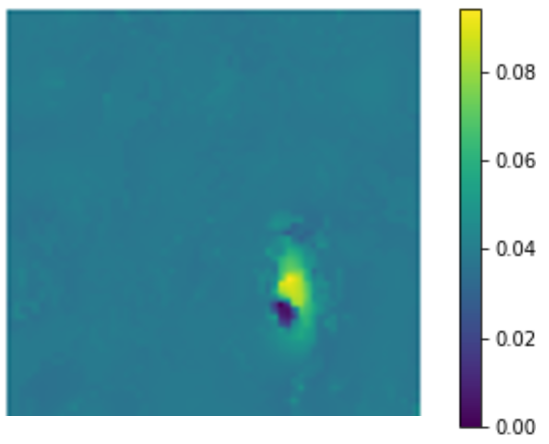}}
\hfill
\subfigure[reconstruction by RL-trained strategy (PSNR$=111.73$)]{\label{fig:k}\includegraphics[width=46mm]{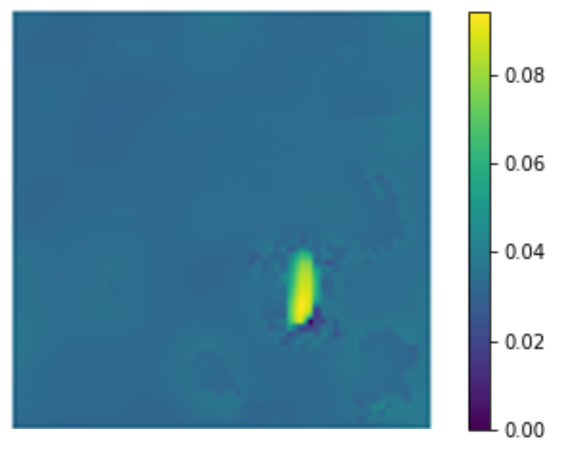}}
\subfigure[Scatterer $4$ Groundtruth]{\label{fig:l}\includegraphics[width=45mm]{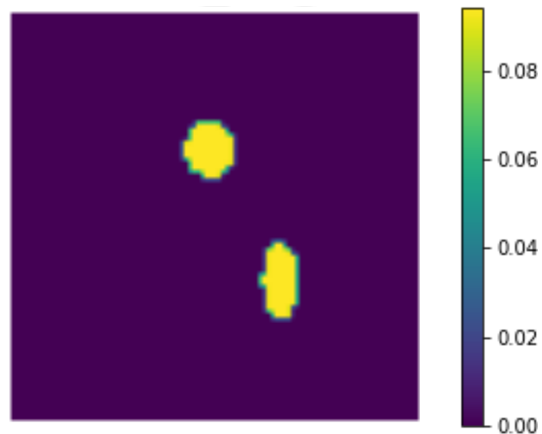}}
\hfill
\subfigure[Reconstruction by the pretrained strategy (PSNR$=112.52$)]{\label{fig:m}\includegraphics[width=45mm]{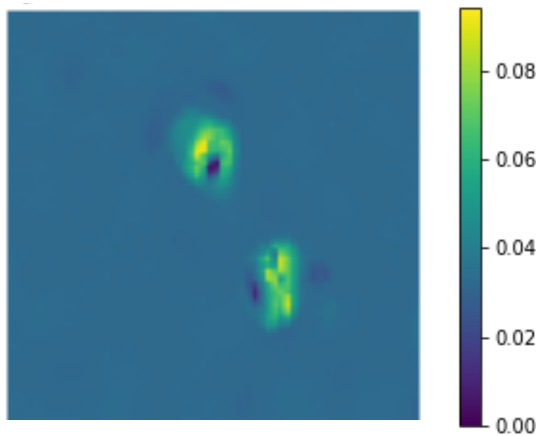}}
\hfill
\subfigure[reconstruction by RL-trained strategy (PSNR$=114.63$)]{\label{fig:n}\includegraphics[width=46mm]{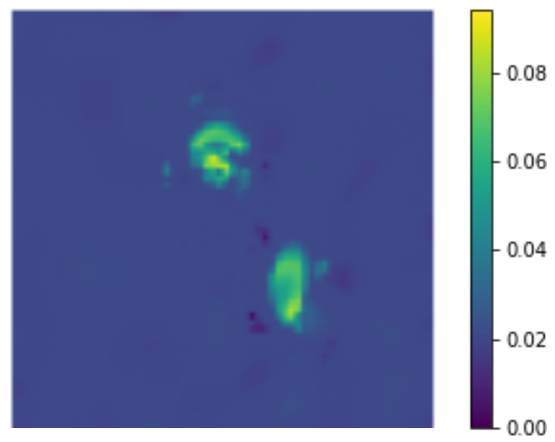}}
\caption{We compare the reconstruction results of the initial strategy and the trained strategy on two specific types of scatterers with sizes of $64\times 64$. The true scatterer is shown in subplots (a), (d), (g) and (j). Each plot is tagged with the respective method, and the PSNR of the reconstruction indicates the difference in resolution.
}
\label{exp64}
\end{figure}

We initially compare the methods on two types of scatterers with a resolution of \(32 \times 32\). The first type is generated by randomly placing three shapes within the unit square. The second type is generated using \(\sin(x+y) f(x,y)\), where \(f(x,y)\) represents a square area containing rectangles. Subsequently, we present the reconstruction results for scatterers with a resolution of \(64 \times 64\) in Figure \ref{exp64}.

\begin{figure}
\centering
\subfigure[Type $1$ scatterer]{\label{fig:a}\includegraphics[width=45mm]{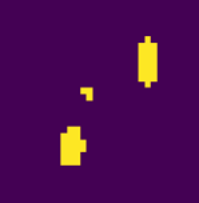}}
\hspace{10mm}
\subfigure[Type $2$ scatterer]{\label{fig:b}\includegraphics[width=45mm]{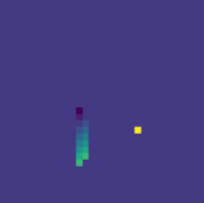}}
\label{example}
\end{figure}

\subsection{Numerical Results}
\label{subsec4.2}
In conclusion, the RL model that learns both angles and frequencies improves significantly compared to other methods under limited sensing resources. We now explain and present the numerical results of several datasets in detail.

\noindent{\bf Strategy 1.} (Uniform angles with a fixed frequency) This strategy uniformly places sensors on the unit circle and uses waves with a fixed frequency, a common approach in the literature. We first train the model to learn this specific policy and then use reinforcement learning (RL) to update the pre-trained model. After updating, the RL strategy selects a wavenumber \(k = 16\), while other methods that do not learn frequencies are assigned a fixed wavenumber of \(k = 32\).

\begin{figure}
\centering
\setcounter{subfigure}{0}
\subfigure[Scatterer $1$ Groundtruth]{\label{fig:a}\includegraphics[width=45mm]{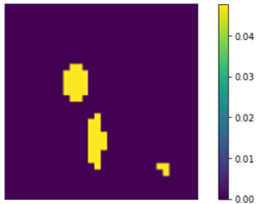}}
\hfill
\subfigure[Reconstruction by the pretrained strategy (PSNR$=122.23$)]{\label{fig:b}\includegraphics[width=43mm]{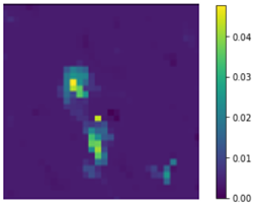}}
\hfill
\subfigure[reconstruction by RL-trained strategy (PSNR$=129.07$)]{\label{fig:c}\includegraphics[width=45mm]{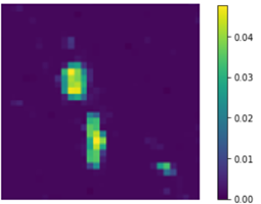}}
\subfigure[Scatterer $2$ Groundtruth]{\label{fig:c}\includegraphics[width=45mm]{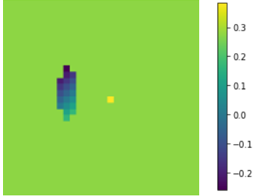}}
\hfill
\subfigure[Reconstruction by the pretrained strategy (PSNR$=103.3$)]{\label{fig:d}\includegraphics[width=45mm]{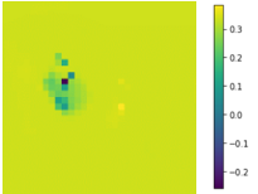}}
\hfill
\subfigure[reconstruction by RL-trained strategy (PSNR$=109.7$)]{\label{fig:e}\includegraphics[width=45mm]{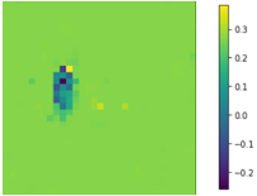}}
\caption{We compare the reconstruction results of the initial strategy and the trained strategy on two specific types of scatterers with sizes of \(32 \times 32\). The true scatterer is shown in subplots (a) and (d). Each plot is tagged with the respective method, and the PSNR of the reconstruction indicates the difference in resolution.
}
\label{exp1f}
\end{figure}

We compare the MSE and PSNR of the methods over $50$ test samples randomly selected from the test set. The reconstructions from our model show a higher PSNR value in more than $90\%$ of the cases. The mean and standard deviation of MSE and PSNR are presented in Tables \ref{table1} and \ref{table2}. The reconstruction results are visualized in Figure \ref{exp1f}. Learning both angles and frequencies results in a significantly smaller MSE and a larger PSNR compared to random or uniform angles. Moreover, learning both angles and frequencies outperforms learning angles only or frequencies only. Our results demonstrate the effectiveness and necessity of training both angles and frequencies. Additionally, the significantly lower variance in MSE indicates better stability of our algorithm compared to others.

\begin{table}[H]
    \centering
    \begin{minipage}{0.45\textwidth}  
        \centering
        \begin{tabular}{llllll}
            \toprule
            & Strategy $1$ & RL-trained Strategy \\      
            \midrule
            MSE  & 7.6e-6  & 2.3e-6   \\ 
            PSNR  & 123.4   & 128.6  \\
            \bottomrule
        \end{tabular}
        \caption{The MSE and PSNR of strategy $1$ and the corresponding trained strategy on scatterer type $1$.}
        \label{table1}
    \end{minipage}%
    \hspace{0.05\textwidth} 
    \begin{minipage}{0.45\textwidth}  
        \centering
        \begin{tabular}{llllll}
            \toprule
            & Strategy $1$ & RL-trained Strategy \\      
            \midrule
            MSE  & 7.1e-4  & 2.1e-4   \\ 
            PSNR  & 103.7   & 108.5  \\
            \bottomrule
        \end{tabular}
        \caption{The MSE and PSNR of strategy $1$ and the corresponding trained strategy on scatterer type $2$.}
        \label{table2}
    \end{minipage}
\end{table}

\begin{figure}
    \centering
    \includegraphics[scale=0.6]{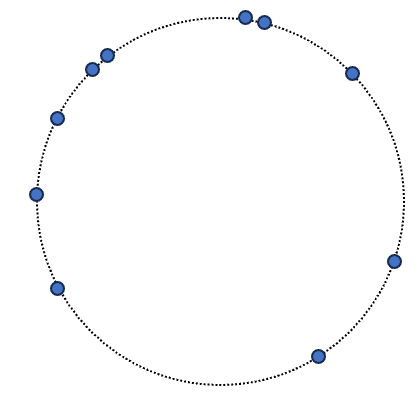}
    \label{pos}
    \caption{The RL-learnt angles for sensors.}
\end{figure}

\noindent{\bf Strategy $2$.}(Random angles with fixed frequency) Strategy $2$ is the second most commonly used approach in the existing literature on inverse scattering. The training scheme for this strategy is identical to the previous one. After training, the RL strategy selects a wavenumber \(k = 32\), while Strategy $2$ is assigned a fixed wavenumber of \(k = 32\). We computed the MSE and PSNR of the methods over $50$ test samples randomly selected from the test set. The results are similar to those of the previous strategy, as shown in Tables \ref{table3} and \ref{table4}.

\begin{table}[H]
    \centering
    \begin{minipage}{0.45\textwidth}  
        \centering
        \begin{tabular}{llllll}
            \toprule
            & Strategy $2$ & RL-trained Strategy \\      
            \midrule
            MSE  & 7.8e-6  & 2.2e-6   \\ 
            PSNR  & 123.2   & 128.9  \\
            \bottomrule
        \end{tabular}
        \caption{The MSE and PSNR of strategy $2$ and the corresponding trained strategy on scatterer type $1$.}
        \label{table3}
    \end{minipage}%
    \hspace{0.05\textwidth} 
    \begin{minipage}{0.45\textwidth}  
        \centering
        \begin{tabular}{llllll}
            \toprule
            & Strategy $2$ & RL-trained Strategy \\      
            \midrule
            MSE  & 8.8e-4  & 2e-4   \\ 
            PSNR  & 102.8   & 109.1  \\
            \bottomrule
        \end{tabular}
        \caption{The MSE and PSNR of strategy $2$ and the corresponding trained strategy on scatterer type $2$.}
        \label{table4}
    \end{minipage}
\end{table}

\noindent{\bf Strategy 3.} (Nonuniform angles with different frequencies) This strategy fixes the angles at \(0, \frac{\pi}{3}, \frac{4\pi}{9}, \frac{5\pi}{9}, \frac{2\pi}{3}, \pi, \frac{4\pi}{3}, \frac{13\pi}{9}, \frac{14\pi}{9}, \frac{5\pi}{3}\). It uses a wavenumber of $8$ at angles \(0\) and \(\pi\), and a wavenumber of $32$ at the other angles. The rationale is to use a higher frequency in the middle, where there are more probes, to achieve higher resolution. The training scheme follows the same procedure as before. After full training, the RL strategy selects a wavenumber \(k = 32\). We computed the MSE and PSNR of the methods over $50$ test samples randomly selected from the test set. The results are presented in Tables \ref{table5} and \ref{table6}. Additionally, it is evident that the performance of the trained strategy is influenced by the initial strategy. Strategy $3$ performs worse on type $1$ scatterers but better on type $2$ scatterers compared to the previous two strategies. The RL-trained strategy also performs worse on type $1$ scatterers but better on type $2$ scatterers.

\begin{table}[H]
    \centering
    \begin{minipage}{0.45\textwidth}  
        \centering
        \begin{tabular}{llllll}
            \toprule
            & Strategy $3$ & RL-trained Strategy \\      
            \midrule
            MSE  & 1e-5  & 2.6e-6   \\ 
            PSNR  & 122.6   & 127.8  \\
            \bottomrule
        \end{tabular}
        \caption{The MSE and PSNR of strategy $3$ and the corresponding trained strategy on scatterer type $1$.}
        \label{table5}
    \end{minipage}%
    \hspace{0.05\textwidth} 
    \begin{minipage}{0.45\textwidth}  
        \centering
        \begin{tabular}{llllll}
            \toprule
            & Strategy $3$ & RL-trained Strategy \\      
            \midrule
            MSE  & 2.6e-4  & 1e-4   \\ 
            PSNR  & 107.7   & 112.2  \\
            \bottomrule
        \end{tabular}
        \caption{The MSE and PSNR of strategy $3$ and the corresponding trained strategy on scatterer type $2$.}
        \label{table6}
    \end{minipage}
\end{table}

The three strategies presented above represent common approaches employed in the field of inverse scattering experiments. These strategies serve as starting points and RL demonstrates its capacity to refine and optimize the initial strategies. The result shows the potential of RL as a powerful tool for enhancing and evolving strategies, ultimately leading to the development of superior approaches.

\section{Discussion}
\label{sec5}
In this paper, reinforcement learning is applied to improve a given policy and it learns to select scatterer-dependent sensing angles and frequencies in inverse scattering. The process of sensor installation, information collection, and scatterer reconstruction is reformulated as a Markov decision process and hence, reinforcement learning can help to optimize this process. A recurrent neural network is adopted as the policy network to choose sensor locations and wave frequencies adaptively. The proposed reinforcement learning method learns to make scatterer-dependent decisions from previous imaging results, each of which requires the solution of an expensive optimization problem. To better facilitate convergence and reduce the computational cost of reinforcement learning, a warm-start strategy is used in these optimization problems. Extensive numerical experiments have been conducted using several types of scatterers with the second and third-order approximation in the nonlinear inverse scattering model. These results demonstrate that the proposed method significantly outperforms the common practice of placing sensors uniformly in inverse scattering problems. The case of weak scattering is adopted as a proof of concept. In the future, more advanced learning techniques will be proposed to deal with more challenging cases.  

In this paper, we employ reinforcement learning to enhance an existing policy by selecting scatterer-dependent sensing angles and frequencies in inverse scattering. The process of sensor deployment, data acquisition, and scatterer reconstruction is reformulated as a Markov decision process, thereby enabling reinforcement learning to optimize this procedure. A recurrent neural network is utilized as the policy network to adaptively determine sensor locations and wave frequencies. The proposed reinforcement learning methodology is designed to make scatterer-dependent decisions based on prior imaging results, each necessitating the solution of a computationally intensive optimization problem. To facilitate convergence and reduce computational expenses, a warm-start strategy is designed. Extensive numerical experiments have been conducted with various types of scatterers using second and third-order approximations in the nonlinear inverse scattering model. The results indicate that the learned strategy significantly outperforms the conventional approach of sensor positioning in inverse scattering problems. The scenario of weak scattering is adopted as a proof of concept. Future work will focus on developing more advanced learning techniques to address increasingly complex scenarios.

\section*{Acknowledgments} Y. K. was partially supported by NSF grant DMS-2111563. H. Y. was partially supported by the US National Science Foundation under awards DMS-2244988, DMS-2206333, the Office of Naval Research Award N00014-23-1-2007, and the DARPA D24AP00325-00.  We thank the authors in the seminal work \citep{ct} for sharing their code.

\bibliographystyle{unsrtnat}
\bibliography{reference}  






\end{document}